\documentclass{article}

\usepackage{arxiv}

\usepackage[utf8]{inputenc} 
\usepackage[T1]{fontenc}    
\usepackage{hyperref}       
\usepackage{url}            
\usepackage{booktabs}       
\usepackage{amsfonts}       
\usepackage{nicefrac}       
\usepackage{microtype}      
\usepackage{lipsum}

\usepackage{lineno}
\usepackage{amsmath,amssymb,graphicx,url,bbm}
\usepackage{xspace,stmaryrd}
\usepackage{textcomp}
\usepackage{multirow, subcaption, adjustbox}
\usepackage{xcolor}
\usepackage{comment}
\modulolinenumbers[5]

\newtheorem{theorem}{Theorem}[section]

\newtheorem{remark}{Remark}[section]

\title{Copula-based conformal prediction for Multi-Target Regression}

\author{
  Soundouss Messoudi \\
  HEUDIASYC - UMR CNRS 7253\\
  Universit\'e de Technologie de Compi\`egne\\
  60203 COMPIEGNE - FRANCE \\
  \texttt{soundouss.messoudi@hds.utc.fr} \\
   \And
 S\'ebastien Destercke \\
  HEUDIASYC - UMR CNRS 7253\\
  Universit\'e de Technologie de Compi\`egne\\
  60203 COMPIEGNE - FRANCE \\
  \texttt{sebastien.destercke@hds.utc.fr} \\
  \And
 Sylvain Rousseau \\
  HEUDIASYC - UMR CNRS 7253\\
  Universit\'e de Technologie de Compi\`egne\\
  60203 COMPIEGNE - FRANCE \\
  \texttt{sylvain.rousseau@hds.utc.fr} \\
}

\begin{document}
\maketitle

\begin{abstract}
There are relatively few works dealing with conformal prediction for multi-task learning issues, and this is particularly true for multi-target regression. This paper focuses on the problem of providing valid (i.e., frequency calibrated) multi-variate predictions. To do so, we propose to use copula functions applied to deep neural networks for inductive conformal prediction. We show that the proposed method ensures efficiency and validity for multi-target regression problems on various data sets.
\end{abstract}


\section{Introduction}

The most common supervised task in machine learning is to learn a single-task, single-output prediction model. However, such a setting can be ill-adapted to some problems and applications. 

On the one hand, producing a single output can be undesirable when data is scarce and when producing reliable, possibly set-valued predictions is important (for instance in the medical domain where examples are very hard to collect for specific targets, and where predictions are used for critical decisions). Such an issue can be solved by using conformal prediction approaches~\cite{shafer2008tutorial}. It was initially proposed as a transductive online learning approach to provide set predictions (in the classification case) or interval predictions (in the case of regression) with a statistical guarantee depending on the probability of error tolerated by the user, but was then extended to handle inductive processes~\cite{papadopoulos2002inductive}.  
On the other hand, there are many situations where there are multiple, possibly correlated output variables to predict at once, and it is then natural to try to leverage such correlations to improve predictions. Such learning tasks are commonly called Multi-task in the literature~\cite{caruana1998dozen}.

Most research work on conformal prediction for multi-task learning focuses on the problem of multi-label prediction~\cite{wang2015comparison,wang2020active}, where each task is a binary classification one. Conformal prediction for multi-target regression has been less explored, with only a few studies dealing with it: Kuleshov \emph{et al.}~\cite{kuleshov2018conformal} provide a theoretical framework to use conformal predictors within manifold (e.g., to provide a mono-dimensional embedding of the multi-variate output), while Neeven and Smirnov~\cite{neeven2018conformal} use a straightforward multi-target extension of a conformal single-output $k$-nearest neighbor regressor~\cite{papadopoulos2011regression} to provide weather forecasts. However, this latter essentially verifies validity (i.e., having well-calibrated outputs) for each individual target. Recently, we proposed a simple method to have an approximate validity for the multi-variate prediction~\cite{messoudi2020conformal}, that generally provided overly conservative results.

In this paper, we propose a new conformal prediction method fitted to multi-target regression, that makes use of copulas~\cite{nelsen1999introduction} (a common tool to model dependence between multi-variate random variables) to provide valid multi-variate predictions. The interest of such a framework is that it remains very easy to apply while linking multi-variate conformal predictions to the theoretically sound framework that are copulas. Experiments also show that it works quite well, and allows to improve upon previous heuristics~\cite{messoudi2020conformal}.

Section~\ref{sec:confMTR} provides a general overview of our problem: a brief introduction to conformal prediction and multi-target regression will be presented in Sections~\ref{sec:conform} and ~\ref{sec:mtr}, before raising the problematic of applying conformal prediction to the multi-target regression setting in Section~\ref{sec:cpmtr}. We will then present our setting in Section~\ref{sec:Cop-Conf-MTR}: we will first recall the needed basic principles and theorems of copulas in Section~\ref{sec:copulas}, before detailing our conformal multi-target approach in Section~\ref{sec:conform_multi}. The experiments and their results are described in Section~\ref{sec:expe}.

\section{Inductive conformal prediction (ICP) for Multi-Target Regression}
\label{sec:confMTR}


This section recalls the basics of inductive conformal regression and multi-target regression, before introducing the issues we will tackle in this paper. 

\subsection{Inductive conformal regression}
\label{sec:conform}

In regression tasks, conformal prediction is a method that provides a statistical guarantee to the predictions by giving an interval prediction instead of a point prediction in the regression case. By statistical guarantee, it is meant that the set-valued predictions cover the true value with a given frequency, i.e., they are calibrated. It was first introduced as a transductive online learning approach~\cite{gammerman2013learning} and then adapted to the inductive framework~\cite{papadopoulos2002inductive} where one uses a model induced from training examples to get conformal predictions for the new instances. The two desirable features in conformal regressors are (a) \textit{validity}, i.e. the error rate does not exceed $\epsilon $ for each chosen confidence level $1 - \epsilon $, and (b) \textit{efficiency}, meaning prediction intervals are as small as possible.

Let $\lbag z_1=(x_1, y_1), z_2=(x_2, y_2), \dots , z_{n}=(x_{n}, y_{n})\rbag$ be the successive pairs of an object $x_i \in X$ and its real-valued label $y_i \in \mathbb{R}$, which constitute the observed examples.
Assuming that the underlying random variables are exchangeable (a weaker condition than i.i.d.), we can predict $y_{n+1} \in \mathbb{R}$ for any new object $x_{n+1} \in X$ by following the inductive conformal framework.

The first step consists of splitting the original data set $Z = \lbag z_1, \dots, z_{n} \rbag$ into a \textit{training set} $Z^{tr} = \lbag z_1, \dots, z_l\rbag$ and a \textit{calibration set} $Z^{cal} = \lbag z_{l+1}, \dots, z_{n}\rbag$, with $|Z^{cal}|= n - l$. Then, an \textit{underlying algorithm} is trained on $Z^{tr}$ to obtain the \textit{non-conformity measure} $A_l$, a measure that evaluates  the strangeness of an example compared to other examples of a bag, called the non-conformity score. Hence, we can calculate the non-conformity score ${\alpha}_k$ for an example $z_k$ compared to the other examples in the bag $\lbag z_1, \dots, z_l \rbag$ with ${\alpha}_k = A_l(\lbag z_1, \dots , z_l \rbag , z_k)$.

By computing the non-conformity score ${\alpha}_i$ for each example $z_i$ of $Z^{cal}$ using this equation, we get the sequence ${\alpha}_{l+1}, \ldots , {\alpha}_{n}$. When making a prediction for a new example $x_{n+1}$, we use the underlying algorithm to associate to any possible prediction $\hat{y}$ its non-conformity score ${\alpha}^{\hat{y}}_{n+1}$, and calculate its \textit{p-value} which indicates the proportion of less conforming examples than $z_{n+1}$, with:
\begin{equation}
p(\hat{y}_{n+1}) = \frac{|\{ i = l+1, \dots , n, n+1 : {\alpha}_i \geq {\alpha}^{\hat{y}}_{n+1} \}|}{n - l + 1}.
\label{eqconfo}
\end{equation}

The final step before producing the conformal prediction consists of choosing the \textit{significance level} $\epsilon \in (0, 1)$ to get a prediction set with a \textit{confidence level} of $1 - \epsilon$, which is the statistical guarantee of coverage of the true value $y_{n+1}$ by the interval prediction $\hat{\mathbf{y}}_{n+1}$ such that 
$$\hat{\mathbf{y}}_{n+1}=\{ \hat{y}_{n+1} \in \mathbb{R}: p(\hat{y}_{n+1}) > \epsilon\}.$$

The most basic non-conformity measure in a regression setting is the absolute difference between the actual value $y_i$ and the predicted value $\hat{y}_i$ by the underlying algorithm. The non-conformity score is then calculated as follows:
\begin{equation}
{\alpha}_i = |y_i - \hat{y}_i|.
\label{eqbase}
\end{equation}

The sequence of non-conformity scores ${\alpha}_{l+1}, \ldots , {\alpha}_{n}$ for all examples in $Z^{cal}$ are obtained and sorted in descending order. Then, we compute the index of the $(1-\epsilon)$-percentile non-conformity score ${\alpha}_{s}$, based on the chosen significance level $\epsilon$, such as:
\begin{equation}
    \mathbb{P}(|y_i - \hat{y_i}| \leq \alpha _s ) \geq 1 - \epsilon.
\end{equation}

Finally, the prediction interval for each new example $x_{n+1}$, which covers the true output $y_{n+1}$ with probability $1 - \epsilon$ is calculated as:
\begin{equation}
\hat{\mathbf{y}}_{n+1}=[{\hat{y}}_{n+1} - {\alpha}_{s}, {\hat{y}}_{n+1} + {\alpha}_{s}].
\end{equation}

The drawback of this standard non-conformity measure is that all prediction intervals are equally sized ($2{\alpha}_{s}$) for a given confidence level. Adopting a \textit{normalized} non-conformity measure instead provides personalized individual bounds for each new example by scaling the standard non-conformity measure with ${\sigma}_{i}$, a term that estimates the difficulty of predicting $y_i$. This means that using a \textit{normalized} non-conformity measure gives a smaller prediction interval for ``easy'' examples, and a bigger one for ``hard'' examples. Thus, two distinct examples with the same ${\alpha}_{s}$ calculated by~\eqref{eqbase} will have two different interval predictions depending on their difficulty. In this case, the normalized non-conformity score is as follows:
\begin{equation}
{\alpha}_i = \frac{|y_i - \hat{y}_i|}{{\sigma}_{i}}.
\label{eqnorm}
\end{equation}
Thus, we have:
\begin{equation}\label{eq:epsilon_mono}
    \mathbb{P}\left( \frac{|y_i - \hat{y_i}|}{\sigma _i } \leq \alpha _s \right) \geq 1 - \epsilon,
\end{equation}
which becomes an equality if the method is perfectly calibrated. For a new example $x_{n+1}$, the prediction interval becomes :
\begin{equation}
\hat{\mathbf{y}}_{n+1}=\left[ {\hat{y}}_{n+1} - {\alpha}_{s}{\sigma}_{n+1}, {\hat{y}}_{n+1} + {\alpha}_{s}{\sigma}_{n+1}\right].
\end{equation}

The value ${\sigma}_{i}$ can be defined in various ways. A popular approach proposed by Papadopoulos and Haralambous~\cite{papadopoulos2011reliable} consists of training a small neural network to estimate the error of the underlying algorithm by predicting the value ${\mu}_i = \ln(|y_i - \hat{y}_i|)$. In this case, the non-conformity score is defined as:
\begin{equation}
{\alpha}_i = \frac{|y_i - \hat{y}_i|}{\exp({\mu}_{i})+\beta},
\label{eqnormexp}
\end{equation}
where $\beta \geq 0$ is a sensitivity parameter. With the significance level $\epsilon$, we have:
\begin{equation}
    \mathbb{P}\left( \frac{|y_i - \hat{y_i}|}{\exp({\mu}_{i})+\beta} \leq \alpha _s \right) \geq 1 - \epsilon.
\end{equation}
For a new example $x_{n+1}$, the prediction interval is:
\begin{equation}
\hat{\mathbf{y}}_{n+1}=\left[{\hat{y}}_{n+1} - {\alpha}_{s}(\exp({\mu}_{n+1})+\beta), {\hat{y}}_{n+1} + {\alpha}_{s}(\exp({\mu}_{n+1})+\beta)\right].
\end{equation}

Other approaches use different algorithms to normalize the non-conformity scores, such as regression trees~\cite{johansson2018interpretable} and $k$-nearest neighbors~\cite{papadopoulos2011regression}. Before introducing the problem of multi-target regression, let us first note that, assuming that our method is well-calibrated and that $|y_i - \hat{y_i}|/\sigma _i$ is associated to a random variable $Q$,~\eqref{eq:epsilon_mono} can be rewritten as 
\begin{equation}\label{eq:cumul_mono}\mathbb{P}( Q \leq \alpha _s ) = 1 - \epsilon := F_{Q}(\alpha_s),\end{equation}
which will be instrumental when dealing with copulas and multi-variate outputs later on. Also note that this means that specifying a confidence $\epsilon$ uniquely defines a value $\alpha_s$.

\subsection{Multi-target regression (MTR)}
\label{sec:mtr}

In multi-target regression, the feature space $X$ is the same as in standard regression, but the target space $Y \subset \mathbb{R}^m$ is made of $m$ real-valued targets. This means that observations are i.i.d pairs $(x_i, y_i)$ drawn from a probability distribution on $X \times Y$, where each instance $x_i \in X$ is associated to an $m$ dimensional real-valued target $y_i = (y_i^1, \ldots , y_i^m) \in Y$. The usual objective of multi-target regression is then to learn a predictor $h: X \rightarrow Y$, i.e. to predict multiple outputs based on the input features characterizing the data set, which generalizes standard regression. There are two distinct approaches to treat MTR called \textit{algorithm adaptation} and \textit{problem transformation} methods.

For \textit{algorithm adaptation} approaches, standard single-output regression algorithms are extended to the multi-target regression problem. Many models were adapted to the MTR problem, such as Support Vector Regressors~\cite{sanchez2004svm}, regression trees~\cite{de2002multivariate}, kernel methods~\cite{baldassarre2012multi} and rule ensembles~\cite{aho2009rule}.

In \textit{problem transformation}, one usually decomposes the initial multi-variate problems into several simpler problems, thus allowing the use of standard classification methods without the need for an adaptation that can be tricky or computationally costly. A prototypical example of such a transformation is the chaining method~\cite{spyromitros2016multi}, where one predicts each target sequentially, using the output and predictions of previous targets as inputs for the next one, thus capturing some correlations between the targets. 

As our goal here is not to produce a new MTR method, but rather to propose a flexible means to make their predictions reliable through conformal prediction, we will not make a more detailed review of those methods. The reader interested in different methods can consult for instance~\cite{spyromitros2016multi}. We will now detail how conformal prediction and MTR can be combined. Let us just mention that exploiting the possible relationships allow in general to improve performances of the methods~\cite{ruder2017overview,caruana1993multitask}. 


\subsection{Inductive conformal prediction for Multi-Target Regression}
\label{sec:cpmtr}

As said before, previous studies about conformal MTR focused on providing valid and efficient inferences target-wise~\cite{neeven2018conformal}, thus potentially neglecting the potential advantages of exploiting target relations. Our main goal in this paper is to provide an easy conformal MTR method allowing to do so.


Within the MTR setting, we have a multi-dimensional output $\{ Y^1 , \ldots , Y^m\}$ (we will use superscripts to denote the dimensions, and subscripts to denote sample indices) with $Y^j \in \mathbb{R}, j \in \{ 1, \ldots , m \}$ the different individual real-valued $m$ targets. Let $\underline{\hat{y}}_{n+1}^j,\overline{\hat{y}}_{n+1}^j$ be respectively the lower and upper bounds of the interval predictions given by the non-conformity measure for each target $Y^j$ given a new instance $x_{n+1}$. We define the hyper-rectangle $[\hat{\mathbf{y}}_{n+1}]$ as the following Cartesian product:
\begin{equation}\label{eq:vol_hyper}
    [\hat{\mathbf{y}}_{n+1}]=\times_{j=0}^m [\underline{\hat{y}}_{n+1}^j,\overline{\hat{y}}_{n+1}^j].
\end{equation}

This hyper-rectangle forms the volume $\prod_{j=0}^m (\overline{\hat{y}}_{n+1}^j - \underline{\hat{y}}_{n+1}^j)$ to which a global prediction $y_{n+1}$ of a new example $x_{n+1}$ should belong in order to be valid, i.e. each single prediction $y_{n+1}^j$ for each individual target $Y^j$ should be between the bounds $\underline{\hat{y}}^j_{n+1},\overline{\hat{y}}^j_{n+1}$ of its interval prediction. With this view, the objective of the conformal prediction framework for MTR in the normalized setting is to satisfy a global significance level $\epsilon_g$ required by the user such that:
\begin{equation}
    \mathbb{P}(y_{n+1} \in [\hat{\mathbf{y}}_{n+1}]) \geq 1 - \epsilon_g.
\end{equation}

This probability can also be written as follows:
\begin{gather}
\mathbb{P}(y_{n+1}^1 \in [\underline{y_{n+1}^1}, \overline{y_{n+1}^1}], \ldots , y_{n+1}^m \in [\underline{y_{n+1}^m}, \overline{y_{n+1}^m}]) \nonumber\\
= \mathbb{P}\left(\frac{|y_{n+1}^1 - \hat{y}_{n+1}^1|}{\sigma_{n+1}^1} \leq \alpha^1_s , \ldots , \frac{|y_{n+1}^m - \hat{y}_{n+1}^m|}{\sigma_{n+1}^m} \leq \alpha^m_s \right) \geq 1 - \epsilon_g.
\end{gather}

Thus, we need to find the individual non-conformity scores $\alpha^1_s , \ldots , \alpha^m_s$, defined for instance by target-wise confidence levels $\epsilon_j$, such  that we ensure a global confidence level $1 - \epsilon_g$. Extending~\eqref{eq:cumul_mono} and considering the random variables $Q^j = |y^j - \hat{y}^j|/\sigma ^j$, $j \in \{ 1, \ldots , m \}$, we get:
\begin{equation}
    \mathbb{P}(Q^1 \leq \alpha^1_s , \ldots , Q^m \leq \alpha^m_s ) \geq 1 - \epsilon_g.
    \label{eqprobacop}
\end{equation}
Should we know the joint distribution in~\eqref{eqprobacop}, and therefore the dependence relations between target predictions, it would be relatively easy to get the individual significance levels\footnote{Note that there may be multiple choices for such individual levels. Here we will fix them to be equal for simplicity.} $\epsilon_j$ associated to the individual non-conformity scores $\alpha^j_s$ such that we satisfy the chosen confidence level $1 - \epsilon_g$. Yet, such a joint distribution is usually unknown. The next section proposes a simple and efficient method to do so, leveraging the connection between~\eqref{eqprobacop} and copulas. Before doing that, note again that under the assumption that we are well calibrated, we can transform~\eqref{eqprobacop} into
\begin{equation}
    F(\alpha^1_s , \ldots ,\alpha^m_s ) = 1 - \epsilon_g,
    \label{eq:probacopcum}
\end{equation}
where $F$ denotes here the joint cumulative distribution induced by $\mathbb{P}$.

\section{Copula-based conformal Multi-Target Regression}
\label{sec:Cop-Conf-MTR}

This section introduces our approach to obtain valid or better conformal prediction in the multi-variate regression setting. We first recall some basics of copulas and refer to Nelsen~\cite{nelsen1999introduction} for a full introduction, before detailing how we apply them to conformal approaches. 

\subsection{Overview on copulas}
\label{sec:copulas}

A copula is a mathematical function that can describe the dependence between multiple random variables. The term ``copula'' was first introduced by Sklar~\cite{sklar1959fonctions} in his famous theorem, which is one of the fundamentals of copula theory, now known as Sklar's theorem. However, these tools have already been used before, as for instance in Fr{\'e}chet's paper~\cite{frechet1951tableaux} and H{\"o}ffding's work~\cite{hoffding1940masstabinvariante, hoeffding1941masstabinvariante} (reprinted as~\cite{hoeffding1994scale}). Copulas are popular in the statistical and financial fields~\cite{embrechts2002correlation}, but they are nowadays more and more used in other domains as well, such as hydrology~\cite{favre2004multivariate}, medicine~\cite{nikoloulopoulos2008multivariate}, and machine learning~\cite{liu2019copula}.

Let $\mathbf{Q} = (Q^1, \ldots , Q^m)$ be an $m$-dimensional random vector composed of the random variables $Q^1, \ldots , Q^m$. Let its cumulative distribution function (c.d.f.) be $F = F_Q : \mathbb{R}^m \rightarrow [0, 1]$. This c.d.f. carries two important pieces of information:

\begin{itemize}
    \item The c.d.f. of each random variable $Q^j$ s.t. $F_j(q^j) = \mathbb{P}(Q^j \leq q^j)$, for all $j \in \{1,\ldots m\}.$
    \item The dependence structure between them.
\end{itemize}

The objective of copulas is to isolate the dependence structure from the marginals $Q^j$ by transforming them into uniformly distributed random variables $U^j$ and then expressing the dependence structure between the $U^j$'s. In other words, an $m$-dimensional copula $C: [0, 1]^m \rightarrow [0, 1]$ is a c.d.f. with standard uniform 
marginals. It is characterized by the following properties:

\begin{enumerate}
    \item $C$ is grounded, i.e. if $u^j = 0$ for at least one $j \in \{1,\ldots , m\}$, then $C(u^1, \ldots , u^m) = 0 $.
    \item If all components of $C$ are equal to 1 except $u^j$ for all $u^j \in [0, 1]$ and $j \in \{1,\ldots , m\}$, then $C(1, \ldots, 1, u^j , 1, \ldots, 1) = u^j$.
    \item $C$ is $m$-increasing, i.e., for all $\mathbf{a}, \mathbf{b} \in [0, 1]^m$ with $\mathbf{a} \leq \mathbf{b}$ :
\begin{equation}
  {\Delta}_{(\mathbf{a},\mathbf{b}]}C = \sum_{j \in \{0, 1\}^m} (-1)^{\sum_{k=1}^{m} j_k}C(a_1^{j_1}b_1^{1-j_1},\dots, a_m^{j_m}b_m^{1-j_m}) \geq 0. \nonumber
\end{equation}

\end{enumerate}
The last inequality simply ensures that the copula is a well-defined c.d.f. inducing non-negative probability for every event. The idea of copulas is based on probability and quantile transformations~\cite{mcneil2015quantitative}. Using these latter, we can see that all multivariate distribution functions include copulas and that we can use a mixture of univariate marginal distributions and a suitable copula to produce a multivariate distribution function. This is described in Sklar's theorem~\cite{sklar1959fonctions} as follows:

\begin{theorem}[Sklar's theorem] For any $m$-dimensional cumulative distribution function  (c.d.f.) $F$ with marginal distributions $F_1,\dots, F_m$, there exists a copula $C: [0,1]^m \rightarrow [0,1]$ such
that:
\begin{equation}
    F(\mathbf{q})=F(q^1, \ldots ,q^m) = C(F_1(q^1), \ldots , F_m(q^m)), \quad\mathbf{q} \in \mathbb{R}^m.
\label{eqsklar}
\end{equation}
If $F_j$ is continuous for all $j \in \{1, \ldots , m\}$, then $C$ is unique. 
\end{theorem}

Denoting the pseudo inverse of $F_j$ as $F^{\leftarrow}_j$~\cite{mcneil2015quantitative}, we can get from~\eqref{eqsklar} that
\begin{equation}
    C(\mathbf{u})=C(u^1, \ldots ,u^m) = F(F^{\leftarrow}_1 (u^1), \ldots , F^{\leftarrow}_m (u^m)).
\end{equation}
There are a few noticeable copulas, among which are:
\begin{itemize}
    \item the product copula: $\Pi (\mathbf{u}) = \prod_{j=1}^{m} u^j$;
    \item the Fr{\'e}chet-H{\"o}ffding upper bound copula \footnote{$M$ is a copula for all $m \geq 2$.}: $M(\mathbf{u}) = \min_{1 \leq j \leq m}\{u^j\}$;
    \item the Fr{\'e}chet-H{\"o}ffding lower bound copula \footnote{$W$ is a copula if and only if $m = 2$.}: $W(\mathbf{u}) = \max\{\sum_{j=1}^{m} u^j - m + 1, 0\}$.
\end{itemize}

While the product copula corresponds to classical stochastic independence, the  Fr{\'e}chet-H{\"o}ffding bound copulas play an important role as they correspond to extreme cases of dependence~\cite{schmidt2007coping}. Indeed, any $m$-dimensional copula $C$ is such that  
$W(\mathbf{u}) \leq C(\mathbf{u}) \leq M(\mathbf{u}), \mathbf{u} \in [0, 1]^m.$

Another important class of copulas are so-called Archimedean copulas, which are based on generator functions $\phi$ of specific kinds. More precisely, a continuous, strictly decreasing, convex function  $\phi : [0, 1] \rightarrow [0, \infty]$ satisfying $\phi (1) = 0$ is known as an Archimedean copula generator. It is known as a strict generator if $\phi (0) = \infty$. The generated copula is then given by
\begin{equation}
    C(u^1, \ldots ,u^m) = {\phi}^{[-1]}(\phi (u^1) + \ldots + \phi (u^m)).
\label{eqmultiarchicop}
\end{equation}

Table~\ref{archifamtb} provides examples and details of three one parameter Archimedean copula families~\cite{mcneil2015quantitative}, which are particularly convenient in estimation problems (being based on a single parameter). 

\begin{table}[ht]
\begin{center}
\begin{tabular}{|c|c|c|c|c|c|}
\hline
Family  & Generator $\phi (t)$                                 & $\theta$ range         & Strict          & Lower & Upper \\ \hline
Gumbel~\cite{gumbel1960distributions} & $(- \ln t)^{\theta}$                                  & $\theta \geq 1$         & Yes             & $\Pi$ & $M$   \\
Clayton~\cite{genest1993statistical} & $\frac{1}{\theta}(t^{-\theta } - 1)$                 & $\theta \geq - 1$       & $\theta \geq 0$ & $W$   & $M$   \\
Frank~\cite{frank1979simultaneous}  & $-\ln\left(\frac{e^{- \theta t} - 1}{e^{- \theta} - 1}\right)$ & $\theta \in \mathbb{R}$ & Yes             & $W$   & $M$   \\ \hline
\end{tabular}
\end{center}
\caption{Archimedean copula families.}
\label{archifamtb}
\end{table}

\subsection{Copula-based conformal Multi-Target Regression}
\label{sec:conform_multi}

Let us now revisit our previous problem of finding the significance levels $\epsilon_j$ for each target so that the hyper-rectangle prediction $[\hat{\mathbf y}]$ covers the true value with confidence $1-\epsilon_g$. Let us first consider~\eqref{eq:probacopcum}. Following Sklar's theorem, we have 
\begin{align*}
     F(\alpha^1_s , \ldots ,\alpha^m_s ) & = C(F_1(\alpha^1_s), \ldots , F_m(\alpha^m_s)) & \\ &= C(1-\epsilon^1, \ldots , 1-\epsilon^m)& \\
     &= 1-\epsilon_g & 
\end{align*}
where the second line is obtained from~\eqref{eq:epsilon_mono}. Clearly, if we knew the copula $C$, then we could search for values $\epsilon_j$ providing the desired global confidence. 

A major issue is then to obtain or estimate the copula modelling the dependence structure between the targets and their confidence levels. As copulas are classically estimated from multi-variate observations, a simple means that we will use here is to estimate them from the non-conformity scores generated from the calibration set $Z^{cal}$. Namely, if $\alpha_i^j$ is the non-conformity score corresponding to the $j^{th}$ target of the $z_i$ example of $Z^{cal}$ for $i \in \{l+1, \ldots , n\}$, we simply propose to estimate a copula $C$ from the matrix
\begin{equation}
A = \begin{bmatrix} 
    \alpha_{l+1}^1 & \alpha_{l+1}^2 & \dots \\
    \vdots & \ddots & \\
    \alpha_{n}^1 &        & \alpha_{n}^m 
    \end{bmatrix}.
\end{equation}

\subsection{On three specific copulas}
\label{sec:3-cop}

We will now provide some detail about the copulas we performed experiments on. They have been chosen to go from the one requiring the most assumptions to the one requiring the least assumptions. 

\subsubsection{The Independent copula}

The Independent copula means that the $m$ targets are considered as being independent, with no relationship between them. It is a strong assumption, but it does not require any estimation of the copula. In this case, \eqref{eqprobacop} becomes:
\begin{align}
\Pi(F_1(\alpha ^1_s), \ldots , F_m(\alpha ^m_s)) 
&= \prod_{j=1}^{m} F_j(\alpha ^j_s) = \prod_{j=1}^{m} \mathbb{P}(Q^j \leq \alpha ^j_s) \nonumber\\
&\geq \prod_{j=1}^{m} (1 - \epsilon ^j) = 1 - \epsilon_g, \nonumber
\end{align}
If we assume that all $\epsilon ^1 , \ldots , \epsilon ^m$ equal the same value $\epsilon_t$, then:
\begin{equation}
\prod_{j=1}^{m} (1 - \epsilon ^j) = (1 - \epsilon_t)^m = 1 - \epsilon_g. \nonumber
\end{equation}
Thus, we simply obtain
\begin{equation}
\epsilon_t = 1 - \sqrt[m]{1 - \epsilon_g}.
\label{eqcorrepsilon}
\end{equation}
This individual significance level $\epsilon_t$ is then used to calculate the different non-conformity scores $\alpha ^j_s$ for each target in the multi-target regression problem for the Independent copula.

\subsubsection{The Gumbel copula}

The Gumbel copula is a member of the Archimedean copula family which depends on only one parameter, and in this sense is a good representative of parametric copulas. It comes down to applying the generator function $\phi (F_j(\alpha^j_s)) = (- \ln F_j(\alpha^j_s))^{\theta}$ and its inverse ${\phi}^{[-1]} (F_j(\alpha ^j_s)) = \exp{-(F_j(\alpha ^j_s))^{1/ \theta}}$ to~\eqref{eqmultiarchicop}, resulting in the expression
\begin{equation}
    C^\theta_G(F_1(\alpha^1_s), \ldots , F_m(\alpha^m_s)) = \exp{-\left(\sum _{j = 1}^{m} \left(- \ln F_j(\alpha^j_s)\right)^{\theta}\right)^{1/ \theta}}.
\label{eqgumbel}
\end{equation}
In this case, we need to estimate the parameter $\theta$. Since the marginals $F_j(\alpha^j)$ are unknown, we also need to estimate them. In our case, we will simply use the empirical c.d.f. induced by the non-conformity scores $\alpha_i^j$ of matrix $A$. An alternative would be to also assume a parametric form of the $F_j$, but this seems in contradiction with the very spirit of non-conformity scores. In particular, we will denote by $\hat{F}_j$ the empirical cumulative distribution such that
\begin{equation*}
    \hat{F}_j(\beta)=\frac{|\{\alpha^j_i:\alpha^j_i \leq \beta, i \in \{l+1,\ldots,n\}\}|}{n-l}, \quad \beta \in \mathbb{R}.
\end{equation*}


The parameter $\theta$ can then be estimated from matrix $A$ using the Maximum Pseudo-Likelihood Estimator~\cite{hofert2019elements} with a numerical optimization, for instance by using the Python library ``copulae''\footnote{https://pypi.org/project/copulae/}. Once this is obtained, we then get for a particular choice of $\epsilon_j$ that 
\begin{align}\label{eqgumbel2}
  C_G^{\hat{\theta}} & = \exp{-\left(\sum _{j=1}^m \left(- \ln (1-\epsilon_j)\right)^{\hat{\theta}}\right)}^{1/ {\hat{\theta}}} \\
                     & = \exp{-\left(\sum _{j=1}^m \left(- \ln F_j(\alpha^j_s)\right)^{\hat{\theta}}\right)}^{1/{\hat{\theta}}}
\end{align}

And we can search for values $\epsilon_j$ that will make this equation equal to $1-\epsilon_g$, using the estimations $\hat{F}_j$. The solution is especially easy to obtain analytically if we consider that $\epsilon^1=\ldots=\epsilon^m=\epsilon_t$, as we then have that 
$$\epsilon_t= 1 - (1 - \epsilon_g)^{1/\sqrt[\theta]{m}},$$ and one can then obtain the corresponding non-conformity scores $\alpha^1_s , \ldots , \alpha^m_s$ by replacing $F_j$ by $\hat{F}_j$.

We chose this particular family of Archimedean copulas because its lower bound is the Independent copula (as seen in Table~\ref{archifamtb}). We can easily verify this by taking $\hat{\theta} = 1$. Thus, we can capture independence if it is verified, and otherwise search in the direction of positive dependence. One reason for such a choice is that previous experiments~\cite{messoudi2020conformal} indicate that the product copula gives overly conservative results. 

\subsubsection{The Empirical copula}

Parametric copulas, as all parametric models, have the advantage of requiring less data to be well estimated, while having the possibly important disadvantage that they induce some bias in the estimation, that is likely to grow as the number of target increases. The Empirical copula presents a non-parametric way of estimating the marginals directly from the observations~\cite{ruschendorf1976asymptotic,ruymgaart1978asymptotic}. It is defined as follows~\cite{hofert2019elements}:
\begin{equation}
    C_{E}(\mathbf{u}) = \frac{1}{n-l}\sum _{i=l+1}^{n}\mathbbm{1}_{\mathbf{u}_{i}\leq \mathbf{u}} = \frac{1}{n-l}\sum _{i=l+1}^{n} \prod _{j=1}^{m}\mathbbm{1}_{u_{i}^j\leq u^j}, \quad\mathbf{u} \in [0, 1]^m,
\label{eqempcop}
\end{equation}
where $\mathbbm{1}_A$ is the indicator function of event $A$, and the inequalities $\mathbf{u}_{i}\leq \mathbf{u}$ for $i \in \{l+1, \ldots , n\}$ need to be understood component-wise. $\mathbf{u}_{i}$ are the pseudo-observations that replace the unknown marginal distributions, which are defined as:
\begin{equation}
    \mathbf{u}_{i} = (u_{i}^1, \ldots , u_{i}^m) = (\hat{F}_{1}(\alpha_{i}^1), \ldots , \hat{F}_{m}(\alpha_{i}^m)), \quad i \in \{l+1, \ldots , n\},
\label{pseudoobs}
\end{equation}
where distributions $\hat{F}_j$ are defined as before. Simply put, the Empirical copula corresponds to consider as our joint probability the Empirical joint cumulative distribution. We then have that 
\begin{equation}
    C_E(F_1(\alpha^1_s), \ldots , F_m(\alpha^m_s)) = \frac{1}{n-l}\sum _{i=l+1}^{n} \prod _{j=1}^{m}\mathbbm{1}_{u_{i}^j\leq F_j(\alpha^j_s)}.
\label{empcopexp}
\end{equation}
Using that $F_j(\alpha^j_s)=1-\epsilon_j$, we can then search for values of $\epsilon_j$, $j=1,\ldots,m$ that will make~\eqref{empcopexp} equal to $1-\epsilon_g$. Note that in this case, even assuming that $\epsilon^1=\ldots=\epsilon^m=\epsilon_t$ will require an algorithmic search, which is however easy as $C_E$ is an increasing function, meaning that we can use a simple dichotomic search.

\section{Evaluation}
\label{sec:expe}

In this section, we describe the experimental setting (underlying algorithm, data sets and performance metrics) and the results of our study.

\subsection{Experimental setting}

We choose to work with a deep neural network as the underlying algorithm. We keep the same underlying algorithm for all non-conformity measures, since our focus is to compare between the three copula functions chosen to get the different non-conformity scores. 

To compute the non-conformity scores over the calibration set, we use the normalized non-conformity score given by~\eqref{eqnormexp} as described in~\cite{papadopoulos2011reliable}, and predict ${\mu}_i = \ln(|y_i - \hat{y}_i|)$ simultaneously for all targets by a single multivariate multi-layer perceptron.  In this case, ${\mu}_i $ represents the estimation of the underlying algorithm's error.
As mentioned before, the approach can be adapted to any conformal regression approach.

Experiments are conducted on normalized data with a mean of 0 and a standard deviation of 1 to simplify the deep neural network optimization, with a 10-fold cross validation to avoid the impact of biased results, and with a calibration set equal to $10\%$ of the training examples for all data sets. We take the value $\beta = 0.1$ for the sensitivity parameter and do not optimize it when calculating the normalizing coefficient ${\mu}_i$. After getting the proper training data $(X^{tr}, Y^{tr})$, calibration data $(X^{cal}, Y^{cal})$ and test data $(X^{ts}, Y^{ts})$ for each fold, we follow the steps described below:

\begin{enumerate}
    \item Train the underlying algorithm (a deep neural network) on the proper training data $(X^{tr}, Y^{tr})$. Its architecture is composed of a first dense layer applied to the input with ``selu'' activation (scaled exponential linear units~\cite{klambauer2017self}), three hidden dense layers with dropouts and ``selu'' activation, and a final dense layer with $m$ outputs and a linear activation.
    \item Predict $\hat{Y}^{cal}$ and $\hat{Y}^{ts}$ for calibration and test data respectively using the underlying algorithm.
    \item Train the normalizing multi-layer perceptron on the proper training data $(X^{tr}, \mu_{tr} = \ln (|Y^{tr} - \hat{Y}^{tr}|)$, corresponding to the error estimation of the underlying algorithm. The normalizing MLP consists of three hidden dense layers with ``selu'' activation and dropouts and a final dense layer with $m$ outputs for predicting all targets simultaneously.
    \item Predict $\mu_{cal}$ and $\mu_{ts}$ for calibration and test data respectively using the normalizing MLP.
    \item If needed, get an estimation\footnote{In the case of the Gumbel copula, we use a Maximum Pseudo-Likelihood Estimator with a numerical optimization using the BFGS algorithm} of the copula $C$ from the matrix $A$ of calibration non-conformity scores.
    \item For each global significance level $\epsilon_g$:
    \begin{itemize}
        \item Get the individual significance level $\epsilon_j = \epsilon_t$ for $j \in \{ 1, \ldots , m\}$ and calculate $\alpha _s = \{\alpha^1_s , \ldots , \alpha^m_s\}$ for all targets using calibration data, according to the methods mentioned in Section~\ref{sec:3-cop}.
        \item Get the interval predictions for the test data with:
        \begin{equation}
        \left[{\hat{Y}}^{ts} - {\alpha}_{s}(\exp({\mu}_{ts})+\beta), {\hat{Y}}^{ts} + {\alpha}_{s}(\exp({\mu}_{ts})+\beta)\right].
        \end{equation}
    \end{itemize}
\end{enumerate}

\begin{remark}\label{rem:ident_conf}We choose $\epsilon_j = \epsilon_t$ for $j \in \{ 1, \ldots , m\}$ as we have no indication that individual targets should be treated with different degree of cautiousness. However, since copulas are functions from $[0,1]^m$ to $[0,1]$, there is in principle no problem in considering different confidence degrees for different tasks, if an application calls for it. How to determine and elicit such degrees is however, to our knowledge, an open question. 
\end{remark}

The implementation was done using Python and Tensorflow. The copula part of our experiments was based on the book~\cite{hofert2019elements} and the Python library ``copulae''.

\begin{table}
\begin{center}
\begin{tabular}{|l|l|l|l|}
\hline
\textbf{Names} & \textbf{Examples} & \textbf{Features} & \textbf{Targets} \\ \hline
music origin~\cite{zhou2014predicting}   & 1059    & 68      & 2      \\ \hline
indoor loc~\cite{torres2014ujiindoorloc}   & 21049    & 520      & 3      \\ \hline
scpf~\cite{tsoumakas11a}   & 1137    & 23      & 3      \\ \hline
sgemm~\cite{nugteren2015cltune}              & 241600   & 14       & 4    \\ \hline
rf1~\cite{tsoumakas11a}                  & 9125     & 64       & 8        \\ \hline
rf2~\cite{tsoumakas11a}                 & 9125     & 576      & 8         \\ \hline
scm1d~\cite{tsoumakas11a}                & 9803     & 280      & 16       \\ \hline
scm20d~\cite{tsoumakas11a}               & 8966     & 61       & 16       \\ \hline
\end{tabular}
\caption{Information on the used multi-target regression data sets.}
\label{tabledatasets}
\end{center}
\end{table}

We use eight data sets with different numbers of targets and varying sizes. They are summarized in Table~\ref{tabledatasets}.

\subsection{Results}

\begin{figure}
\centering
  \begin{subfigure}[b]{0.49\textwidth}
    \includegraphics[width=\textwidth]{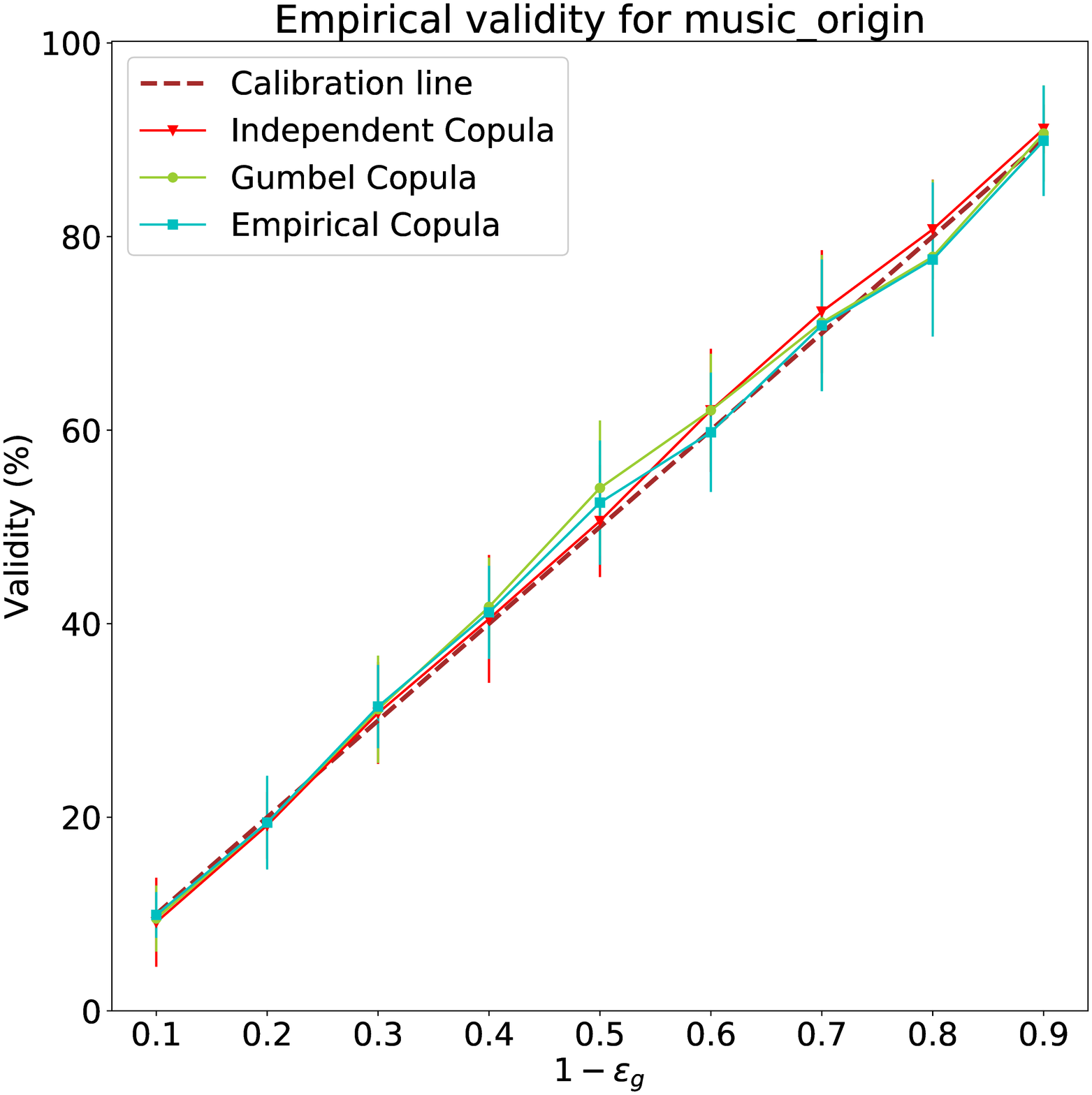}
    \caption{Empirical validity}
    \label{fig:01}
  \end{subfigure}
  \begin{subfigure}[b]{0.49\textwidth}
    \includegraphics[width=\textwidth]{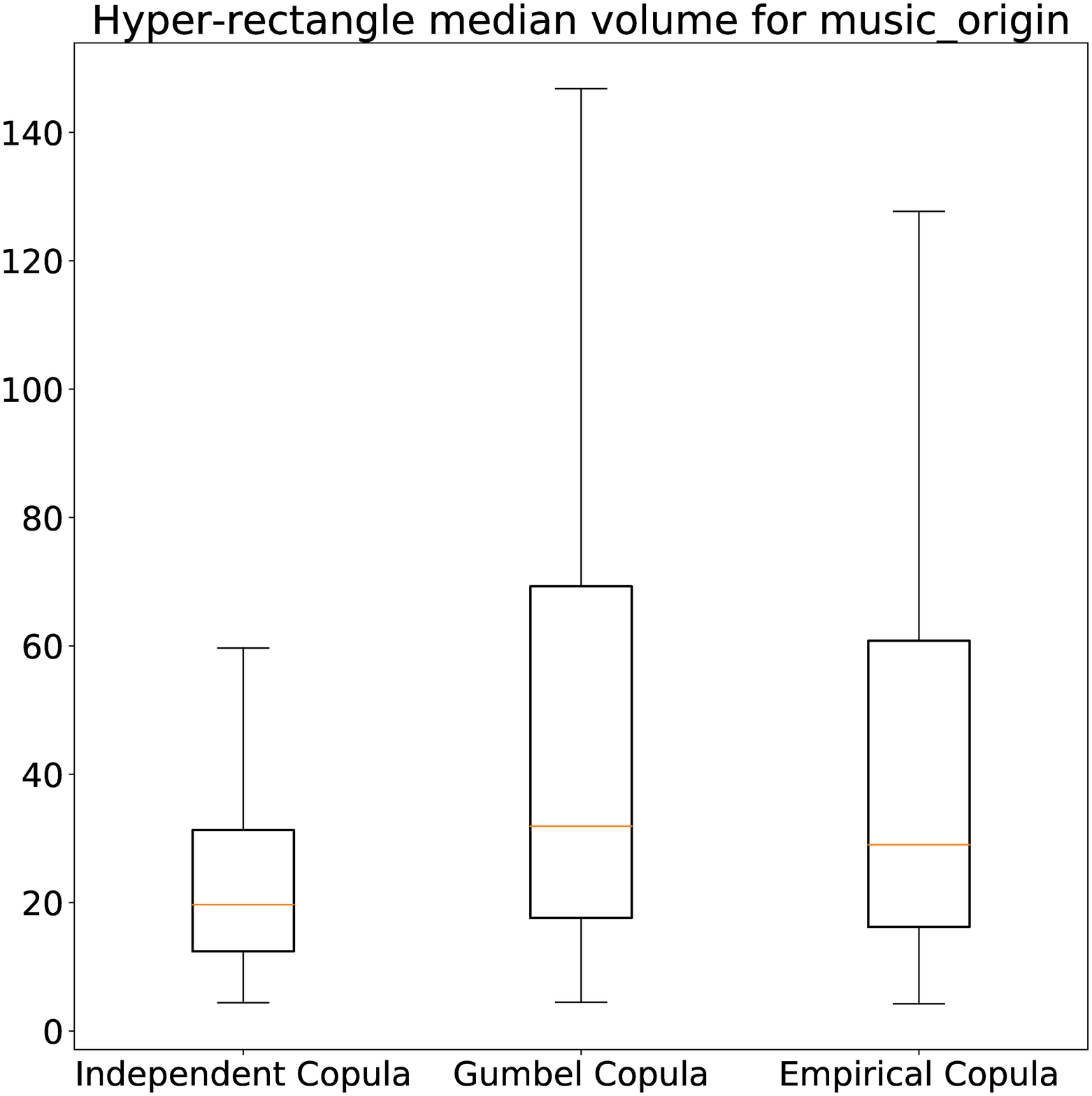}
    \caption{Hyper-rectangle median volume}
    \label{fig:02}
  \end{subfigure}
\caption{Results for music origin.}
\label{fig:fig0}
\end{figure}

\begin{figure}
\centering
  \begin{subfigure}[b]{0.49\textwidth}
    \includegraphics[width=\textwidth]{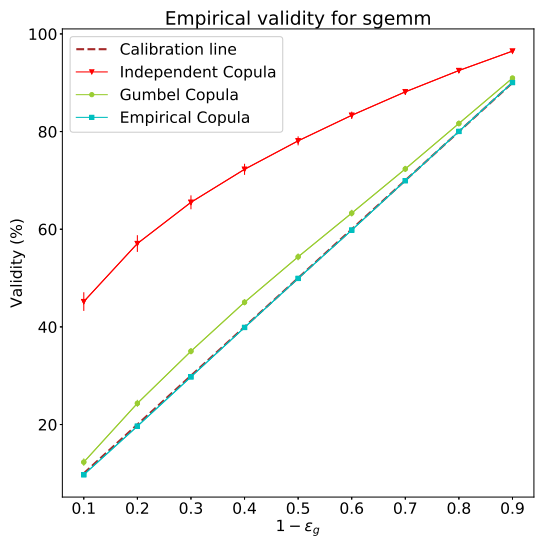}
    \caption{Empirical validity}
    \label{fig:3}
  \end{subfigure}
  \begin{subfigure}[b]{0.49\textwidth}
    \includegraphics[width=\textwidth]{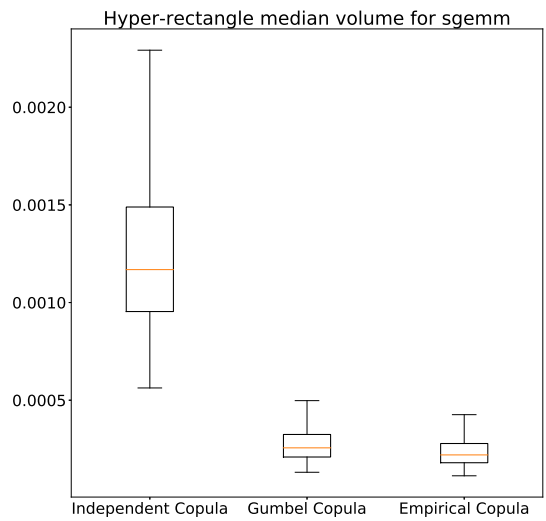}
    \caption{Hyper-rectangle median volume}
    \label{fig:4}
  \end{subfigure}
\caption{Results for sgemm.}
\label{fig:fig2}
\end{figure}

This section presents the results of our experiments, investigating in particular the validity and efficiency of the proposed approaches. Figures~\ref{fig:fig0} and~\ref{fig:fig2} detail these results for ``music origin'' and ``sgemm''. The figures for all other data sets can be found in~\ref{sec:appendix1}.

To verify the validity of each non-conformity measure, we calculate the accuracy of each one and compare it with the calibration line. This line represents the case where the error rate is exactly equal to $\epsilon_g$ for a confidence level $1 - \epsilon_g$, which is the desired outcome of using conformal prediction. In multi-target regression, the accuracy is computed based on whether the observation $y$ belongs to the hyper-rectangle $[\hat{\mathbf{y}}]$ or not depending on the significance level $\epsilon_g$. Thus, a correctly predicted example must verify that all its individual predictions $y_i$ for each individual target $Y_i$ is in its corresponding individual interval predictions. Concretely, for each considered confidence level $\epsilon_g$ and test example $x \in X^{ts}$, we obtain a prediction $[\hat{\mathbf{y}}]_{\epsilon_g}$. From this, we can compute the empirical validity as the percentage of times that $[\hat{\mathbf{y}}]_{\epsilon_g}$ contains the true observed value, i.e., 
$$\frac{\sum_{(x,y) \in Z^{ts}} \mathbbm{1}_{y \in [\hat{\mathbf{y}}]_{\epsilon_g}}}{|Z^{ts}|}.$$
Doing it for several values of $\epsilon_g$, we obtain a calibration curve that should be as close as possible to the identity function.

The results of the error rate or accuracy curves are shown in sub-figure a of each figure for the Independent, Gumbel and Empirical multivariate non-conformity measures. The outcomes clearly show that the best performance is obtained by using the Empirical copula, where the model is well calibrated. For most of the studied data sets, the Empirical copula accuracy curve is almost perfectly aligned with the calibration line, and thus almost exactly valid. This is due to the fact that Empirical copula functions non-parametrically estimate the marginals based on the observations, which enables the model to better adapt to the dependence structure of each data set. This dependence structure is neglected when using an Independent copula-based non-conformity measure, as the $m$ targets are treated as if they were independent, and so the link between them is not exploited when computing $\epsilon_t$. This also means that the difference between the Empirical and the Independent copula-based non-conformity measures is bigger when there is a strong dependence between the non-conformity scores, and is an indication of the strength of this dependence. For instance, we can deduce that the targets are strongly related for ``sgemm'' by the big gap between the Independent and Empirical accuracy curves (sub-figure~\ref{fig:3}). For the Gumbel copula, the accuracy curve is generally closer to the calibration line than the one for the Independent copula. This supports the existence of a dependence structure between the targets, since the lower bound of the Gumbel copula is the Independent copula, which means that if the targets were in fact independent, the two curves would perfectly match. This can be seen in sub-figure~\ref{fig:01} for ``music origin'', where the accuracy curves almost overlap all the time, meaning that the targets are likely to be independent.

From the empirical validity results, we also noticed that the Empirical copula non-conformity measure can be slightly invalid sometimes (sub-figure~\ref{fig:21} for ``scpf''). We explain this by the fewer number of examples, in which case one could use a more regularized form than the Empirical copula. However, when a lot of examples are available (for instance, more than 20000 observations for ``sgemm''), the validity curve of the Empirical copula non-conformity measure is perfectly aligned with the calibration line, meaning that this measure is exactly valid (sub-figure~\ref{fig:3}).

In single-output regression, efficiency is measured by the size of the intervals, and a method is all the more efficient as predicted intervals are small. To assess efficiency in multi-target regression, we can simply compute the volume of the obtained predictions $[\hat{\mathbf{y}}]_{\epsilon_g}$, after~\eqref{eq:vol_hyper}. For each experiment, we then compute the median value of those hyper-rectangle volumes (for the estimation to be robust against very large hyper-rectangles). 

Efficiency results are shown in sub-figure b for all data sets for $\epsilon_g = 0.1$. They show that, in general, the Independent copula has a bigger median hyper-rectangle volume compared to the Gumbel and Empirical copulas, especially in those cases where the existence of a dependence structure is confirmed by the calibration curves. This is due to the fact that using an Independent copula ignores the dependence between the non-conformity scores, which leads to an over-estimation of the global hyper-rectangle error. This impact is avoided when using the Empirical copula because it takes advantage of the dependence structure to construct better interval predictions. Another remark concerning efficiency is that the box plots for Empirical copula are tighter than the other two, which shows that the values are homogeneous on all folds compared to the Independent copula for instance, where the variation is much more visible.

The empirical validity and hyper-rectangle median volume results are summarized in Tables~\ref{tabresval} and~\ref{tabreseff}. The validity simply provides the average difference between a perfect calibration (the identity function) and the observed curve for each copula. This means, in particular, that a negative value indicates that the observed frequency is in average below the specified confidence degree. 

\begin{table}
\begin{center}
\begin{adjustbox}{width={\textwidth},totalheight={\textheight},keepaspectratio}%
\begin{tabular}{|l|c|c|c|}
\hline
Data sets    & Independent                 & Gumbel                         & Empirical                                            \\ \hline
music origin & $7.06 \times 10^1 \pm 5.12$ & $8.48 \times 10^1 \pm 5.72$    & $\mathbf{2.90 \times 10^1 \pm 5.48}$                  \\ \hline
indoor loc   & $2.99 \pm 1.17$             & $2.00 \pm 1.28$                & $\mathbf{3.24 \times 10^{-1} \pm 1.28}$              \\ \hline
scpf         & $9.04 \pm 5.07$             & $2.73 \pm 5.64$                & $\mathbf{-1.42 \pm 4.16}$                             \\ \hline
sgemm        & $2.54 \times 10^1 \pm 1.00$ & $3.26 \pm 6.53 \times 10^{-1}$ & $\mathbf{-1.35 \times 10^1 \pm 3.00 \times 10^{-1}}$  \\ \hline
rf1          & $5.60 \pm 1.59$             & $3.46 \pm 1.56$                & $\mathbf{-9.35 \times 10^{-3} \pm 1.51}$              \\ \hline
rf2          & $6.09 \pm 1.86$             & $2.19 \pm 2.27$                & $\mathbf{-3.61 \times 10^{-1} \pm 2.14}$              \\ \hline
scm1d        & $1.44 \times 10^1 \pm 1.82$ & $1.03 \times 10^1 \pm 2.98$    & $\mathbf{-7.03 \times 10^{-1} \pm 2.32}$             \\ \hline
scm20d       & $1.68 \times 10^1 \pm 1.43$ & $1.02 \times 10^1 \pm 2.35$    & $\mathbf{-1.34 \pm 2.25}$                            \\ \hline
\end{tabular}
\end{adjustbox}
\caption{Validity (average gap between the empirical validity curve and the calibration line in percentage) summarized results for all data sets.}
\label{tabresval}
\end{center}
\end{table}

\begin{table}
\begin{center}
\begin{adjustbox}{width={\textwidth},totalheight={\textheight},keepaspectratio}%
\begin{tabular}{|l|c|c|c|}
\hline
Data sets    & Independent                                   & Gumbel                                        & Empirical                                     \\ \hline
music origin & $\mathbf{1.97 \times 10^1 \pm 2.99}$                   & $3.19 \times 10^1 \pm 1.73 \times 10^1$       & $2.90 \times 10^1 \pm 1.39 \times 10^1$       \\ \hline
indoor loc   & $1.70 \times 10^{-1} \pm 5.12 \times 10^{-2}$ & $9.54 \times 10^{-2} \pm 2.04 \times 10^{-2}$ & $\mathbf{8.69 \times 10^{-2} \pm 1.86 \times 10^{-2}}$ \\ \hline
scpf         & $5.10 \pm 5.31$                               & $3.06 \pm 3.7$                                & $\mathbf{2.39 \pm 3.67}$                               \\ \hline
sgemm        & $1.17 \times 10^{-3} \pm 5.69 \times 10^{-4}$ & $2.56 \times 10^{-4} \pm 1.95 \times 10^{-4}$ & $\mathbf{2.20 \times 10^{-4} \pm 1.60 \times 10^{-4}}$ \\ \hline
rf1          & $1.18 \times 10^{-2} \pm 1.52 \times 10^{-2}$ & $5.52 \times 10^{-3} \pm 1.05 \times 10^{-2}$ & $\mathbf{3.61 \times 10^{-3} \pm 6.00 \times 10^{-3}}$ \\ \hline
rf2          & $2.56 \times 10^{-3} \pm 1.87 \times 10^{-3}$ & $7.48 \times 10^{-4} \pm 8.44 \times 10^{-4}$ & $\mathbf{7.00 \times 10^{-4} \pm 8.48 \times 10^{-4}}$ \\ \hline
scm1d        & $3.49 \times 10^4 \pm 2.89 \times 10^4$       & $1.28 \times 10^4 \pm 1.20 \times 10^4$       & $\mathbf{1.15 \times 10^3 \pm 1.22 \times 10^3}$       \\ \hline
scm20d       & $5.43 \times 10^6 \pm 4.43 \times 10^6$       & $1.80 \times 10^5 \pm 2.15 \times 10^5$       & $\mathbf{4.14 \times 10^4 \pm 6.66 \times 10^4}$       \\ \hline
\end{tabular}
\end{adjustbox}
\caption{Efficiency (hyper-rectangle median volume for $\epsilon_g = 0.1$) summarized results for all data sets.}
\label{tabreseff}
\end{center}
\end{table}

The numbers confirm our previous observations on the graphs, as the average gap is systematically higher for the Independent copula and lower for the Empirical one, with Gumbel in-between. We can however notice that while the Empirical copula provides the best results, it is also often a bit under the calibration line, indicating that if conservativeness is to be sought, one should maybe prefer the Gumbel copula. About the same conclusions can be given regarding efficiency, with the Empirical copula giving the best results and the Independent one the worst. 

\section{Conclusion and discussion}

In this paper, we provided a quite easy and flexible way to obtain valid conformal predictions in a multi-variate regression setting. We did so by exploiting a link between non-conformity scores and copulas, a commonly used tool to model multi-variate distribution. 

Experiments on various data sets for a small choice of representative copulas show that the method indeed allows to improve upon the naive independence assumption. Those first results indicate in particular that while parametric, simple copulas may provide valid results for some data sets, more complex copulas may be needed in general to obtain well calibrated predictions, with the cost that good estimations of such copulas require a lot of calibration data. 

As future lines of work, we would like to explore further the flexibility of our framework, for instance by exploring the possibility of using vines~\cite{joe2011dependence} to model complex dependencies, or by proposing protocols allowing to obtain $\epsilon_g$ from different individual, user-defined confidence degrees, taking up on our Remark~\ref{rem:ident_conf}. 

Finally, while we mostly focused on multi-variate regression in the present paper, it would be interesting to try to extend the current approach to other multi-task settings, such as multi-label problems. A possibility could be to make such problems continuous, as proposed for instance by Liu~\cite{liu2019copula}.

\section{Acknowledgments}

This research was supported by the UTC foundation.

\bibliographystyle{unsrt}
\bibliography{references}

\appendix
\section{Validity and efficiency figures}
\label{sec:appendix1}

This appendix contains the figures for empirical validity and hyper-rectangle median volume for all remaining data sets.

\begin{figure}[ht]
\centering
  \begin{subfigure}[b]{0.49\textwidth}
    \includegraphics[width=\textwidth]{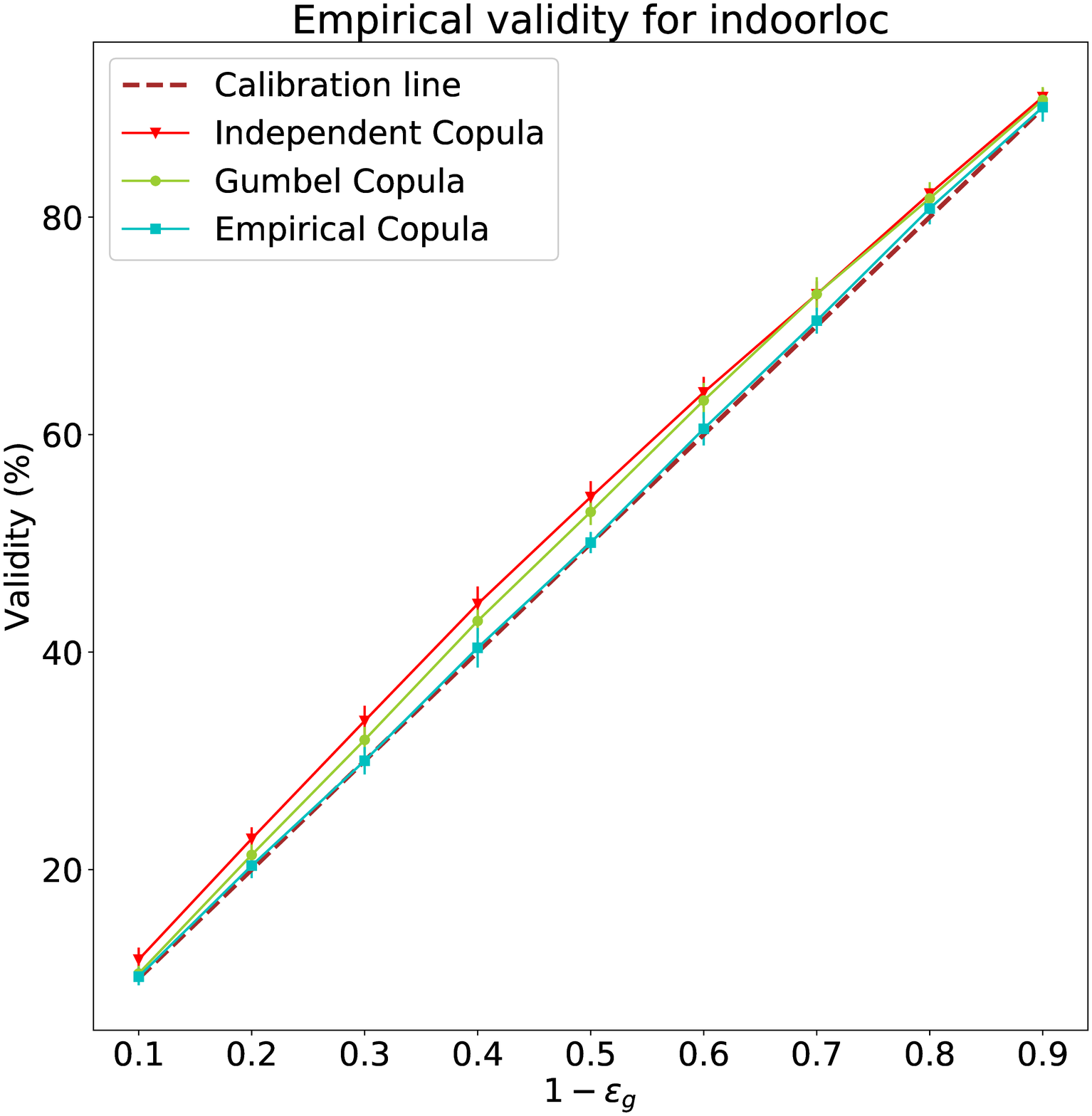}
    \caption{Empirical validity}
    \label{fig:1}
  \end{subfigure}
  \begin{subfigure}[b]{0.49\textwidth}
    \includegraphics[width=\textwidth]{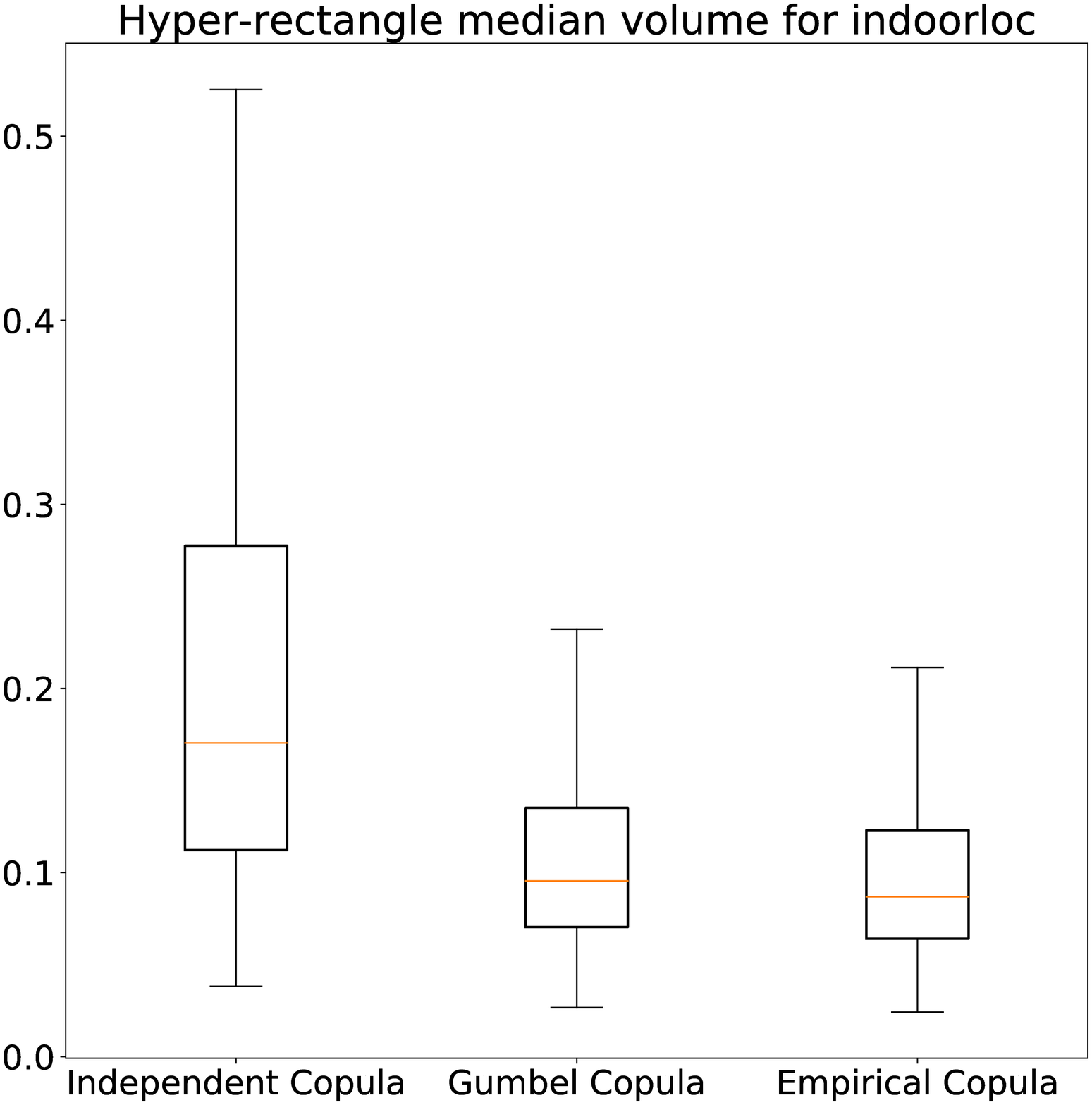}
    \caption{Hyper-rectangle median volume}
    \label{fig:2}
  \end{subfigure}
\caption{Results for indoor loc.}
\label{fig:fig1}
\end{figure}

\begin{figure}[ht]
\centering
  \begin{subfigure}[b]{0.49\textwidth}
    \includegraphics[width=\textwidth]{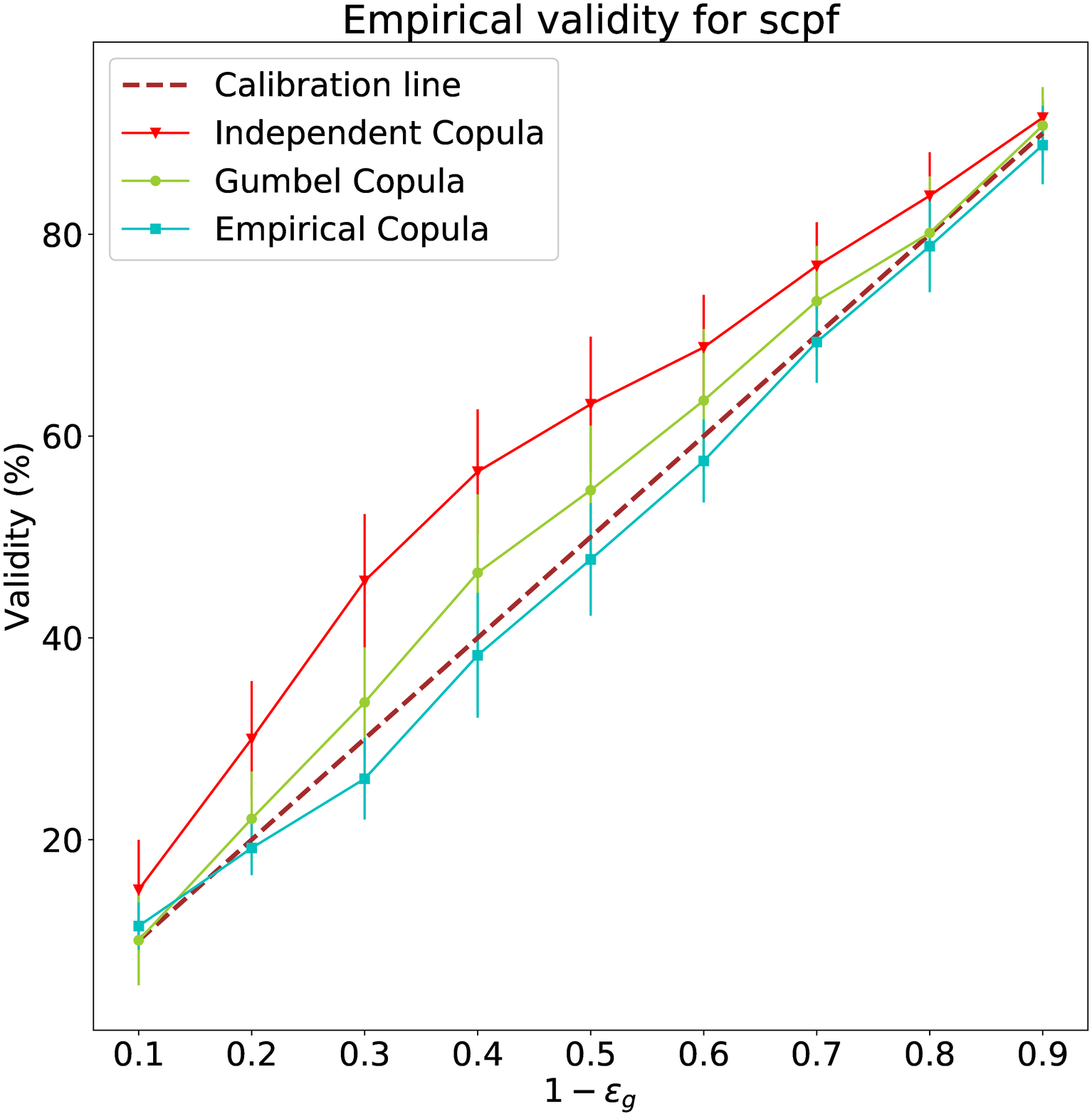}
    \caption{Empirical validity}
    \label{fig:21}
  \end{subfigure}
  \begin{subfigure}[b]{0.49\textwidth}
    \includegraphics[width=\textwidth]{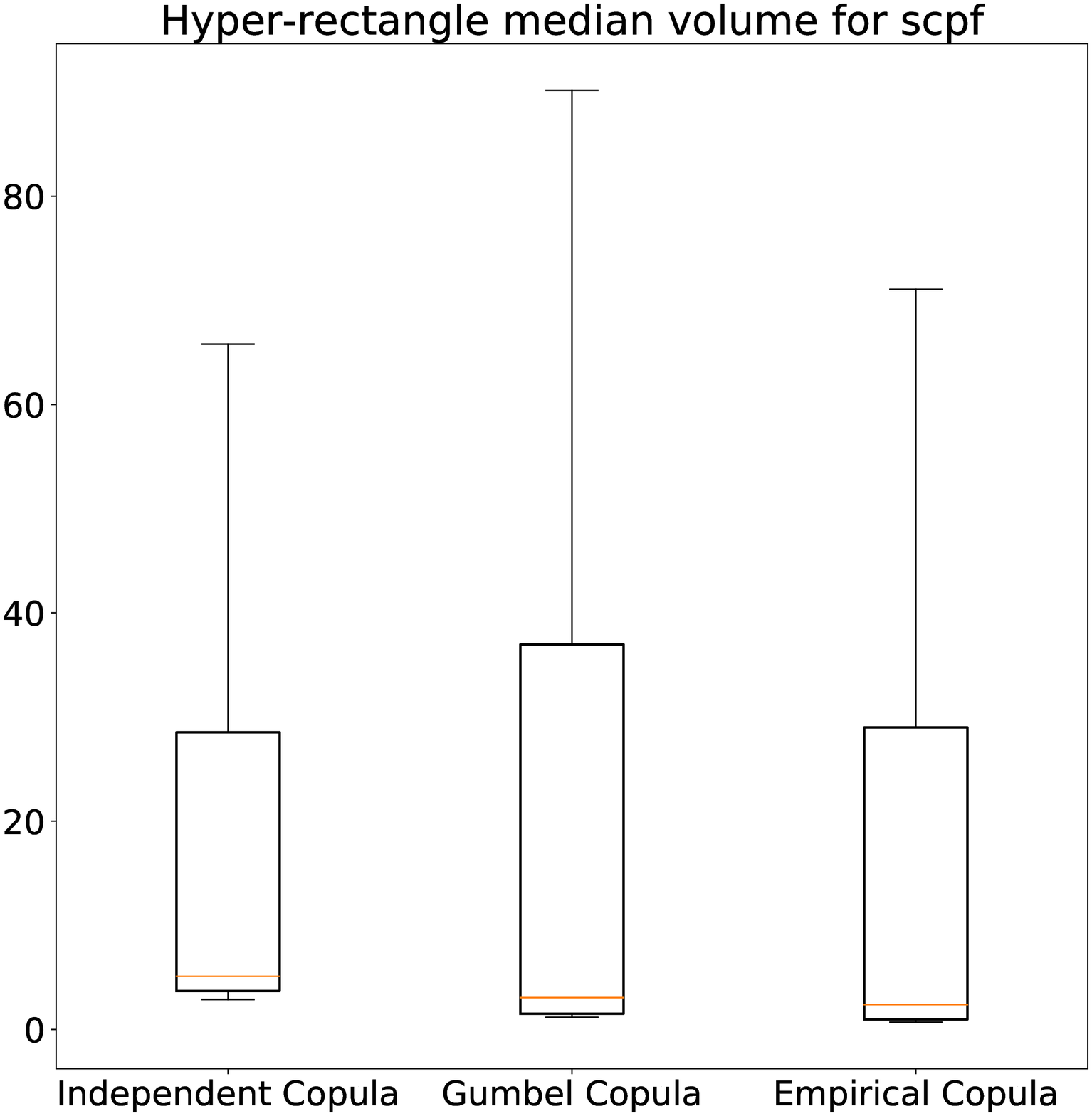}
    \caption{Hyper-rectangle median volume}
    \label{fig:22}
  \end{subfigure}
\caption{Results for scpf.}
\label{fig:fig22}
\end{figure}

\begin{figure}[ht]
\centering
  \begin{subfigure}[b]{0.49\textwidth}
    \includegraphics[width=\textwidth]{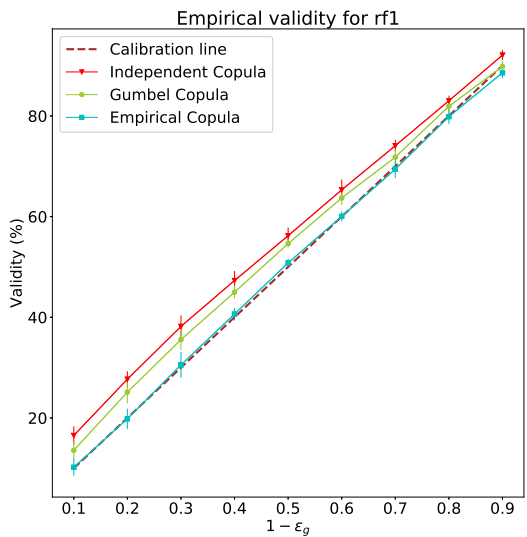}
    \caption{Empirical validity}
    \label{fig:5}
  \end{subfigure}
  \begin{subfigure}[b]{0.49\textwidth}
    \includegraphics[width=\textwidth]{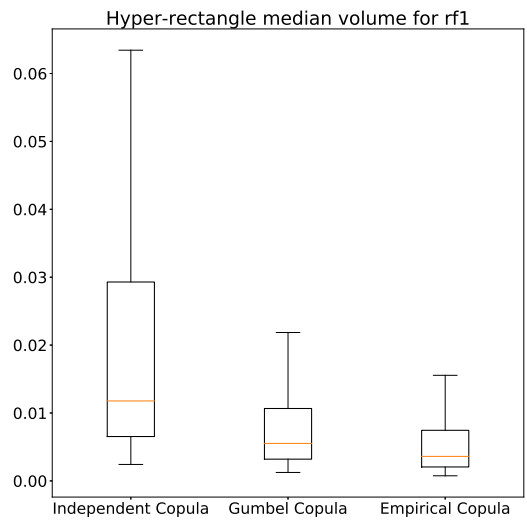}
    \caption{Hyper-rectangle median volume}
    \label{fig:6}
  \end{subfigure}
\caption{Results for rf1.}
\label{fig:fig3}
\end{figure}

\begin{figure}[ht]
\centering
  \begin{subfigure}[b]{0.49\textwidth}
    \includegraphics[width=\textwidth]{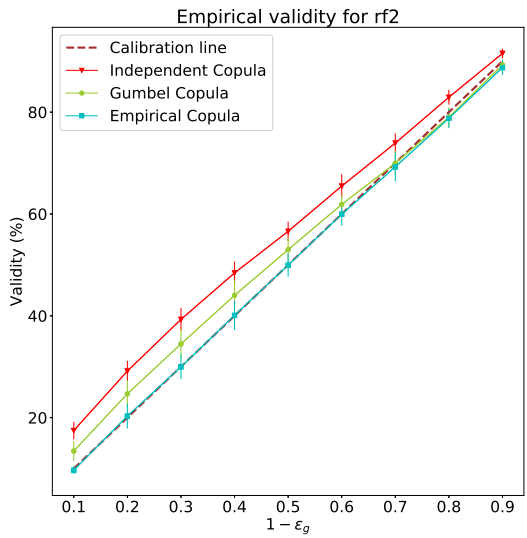}
    \caption{Empirical validity}
    \label{fig:7}
  \end{subfigure}
  \begin{subfigure}[b]{0.49\textwidth}
    \includegraphics[width=\textwidth]{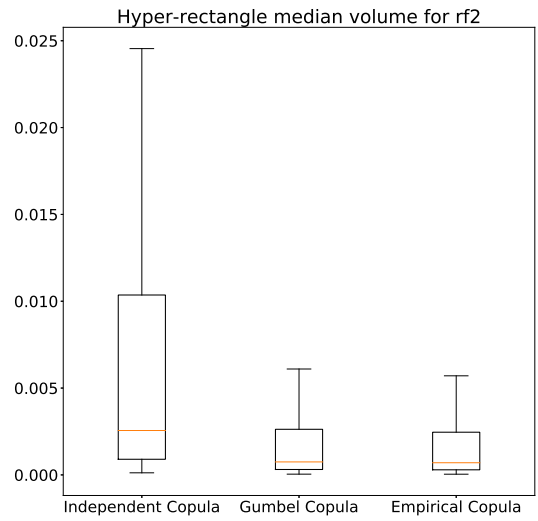}
    \caption{Hyper-rectangle median volume}
    \label{fig:8}
  \end{subfigure}
\caption{Results for rf2.}
\label{fig:fig4}
\end{figure}

\begin{figure}[ht]
\centering
  \begin{subfigure}[b]{0.49\textwidth}
    \includegraphics[width=\textwidth]{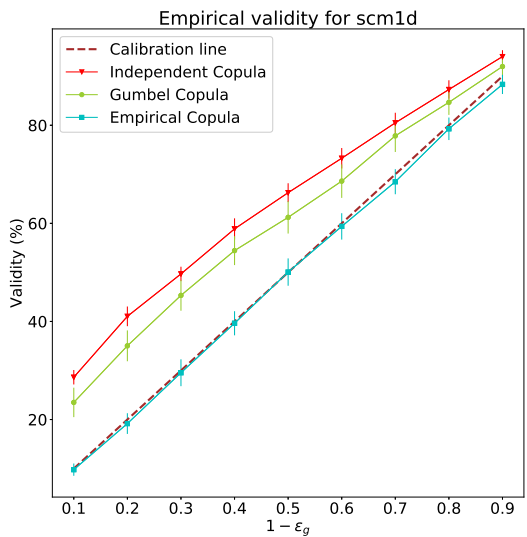}
    \caption{Empirical validity}
    \label{fig:9}
  \end{subfigure}
  \begin{subfigure}[b]{0.49\textwidth}
    \includegraphics[width=\textwidth]{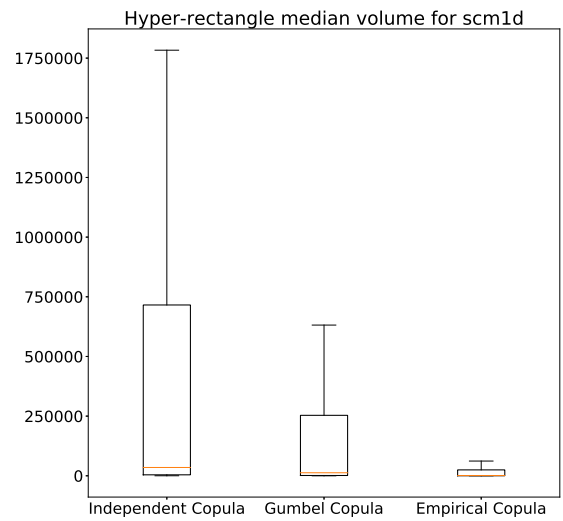}
    \caption{Hyper-rectangle median volume}
    \label{fig:10}
  \end{subfigure}
\caption{Results for scm1d.}
\label{fig:fig5}
\end{figure}

\begin{figure}[ht]
\centering
  \begin{subfigure}[b]{0.49\textwidth}
    \includegraphics[width=\textwidth]{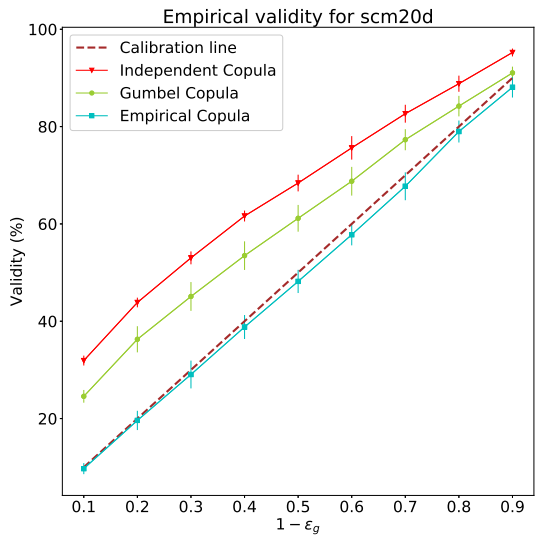}
    \caption{Empirical validity}
    \label{fig:11}
  \end{subfigure}
  \begin{subfigure}[b]{0.49\textwidth}
    \includegraphics[width=\textwidth]{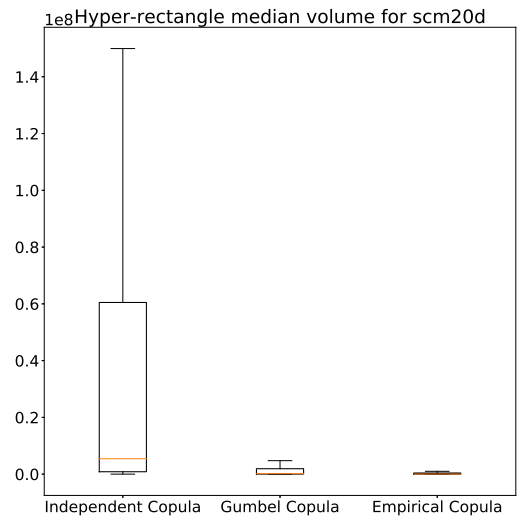}
    \caption{Hyper-rectangle median volume}
    \label{fig:12}
  \end{subfigure}
\caption{Results for scm20d.}
\label{fig:fig6}
\end{figure}

\end{document}